\documentclass[preprint,12pt,authoryear]{elsarticle}
\usepackage{amsmath,amsfonts,amssymb}
\usepackage{array}
\usepackage{textcomp}
\usepackage{url}
\usepackage{graphicx}
\usepackage{booktabs}
\usepackage{multirow}
\usepackage[table]{xcolor}
\usepackage{tikz}
\usetikzlibrary{arrows.meta,positioning,shapes.geometric,calc,fit,backgrounds,shadows.blur,3d}
\usepackage{pgfplots}
\pgfplotsset{compat=1.17}
\usepackage{hyperref}

\newcommand{\networklayer}[7]{%
    \begin{scope}[shift={(#1, -#3/2, -#4/2)}]
        \draw[fill=#5!70!white, draw=black!70, thin, line join=round]
            (0, #3, 0) -- (#2, #3, 0) -- (#2, #3, #4) -- (0, #3, #4) -- cycle;
        \draw[fill=#5!90!black, draw=black!70, thin, line join=round]
            (#2, 0, 0) -- (#2, #3, 0) -- (#2, #3, #4) -- (#2, 0, #4) -- cycle;
        \draw[fill=#5, draw=black!70, thin, line join=round]
            (0, 0, 0) -- (#2, 0, 0) -- (#2, #3, 0) -- (0, #3, 0) -- cycle;
        \ifx&#6& \else \node[anchor=north, font=\scriptsize, align=center] at (#2/2, -0.5, 0) {#6}; \fi
        \ifx&#7& \else \node[anchor=south, font=\scriptsize, align=center, text=black] at (#2/2, #3+0.3, #4) {#7}; \fi
    \end{scope}%
}

\makeatletter
\let\ps@pprintTitle\ps@plain
\makeatother
\begin{document}

\begin{frontmatter}

\title{Now You See Them, Now You Don't: FusionGhost-Gait for Occlusion-Robust Gait Recognition with the ArabGait Dataset 
}

\author[aus]{Seba Al Mokdad\corref{cor1}}
\ead{g00105436@aus.edu}
\ead[url]{https://orcid.org/0000-0003-4938-5493}
\author[aus]{Tamer Shanableh}
\ead{tshanableh@aus.edu}
\author[aus]{Reham Aburas}
\ead{raburas@aus.edu}
\cortext[cor1]{Corresponding author}
\affiliation[aus]{organization={Department of Computer Science and Engineering, American University of Sharjah},
                  city={Sharjah},
                  country={United Arab Emirates}}

\begin{abstract}
Skeleton-based gait recognition is generally more robust to appearance
changes than pixel-based methods, but this robustness depends on stable
pose extraction.
Clothing variation in gait recognition includes changes in garments,
silhouette, and carried items; under Gulf clothing (Kandoora/Abaya), the
dominant failure mode shifts from appearance change to lower-limb pose
degradation caused by severe occlusion.
To study this regime, we introduce ArabGait, a 45-subject indoor
multi-view dataset with four walking conditions across three viewpoints,
and show that Gulf clothing lowers mean lower-limb confidence from 0.811
to 0.683 and raises lower-limb failure rate from 7.98\% to 17.82\%, with
ankles most affected.
We propose FusionGhost-Gait, a fused spatiotemporal architecture that
integrates spatial graph blocks and temporal modeling in one residual unit, removes
redundant channel projections, and uses compact temporal Ghost kernels as
a lightweight regularization mechanism for pose-noisy trajectories while
preserving efficient temporal downsampling.
On ArabGait cross-clothing evaluation, including Protocol A and Protocol
B, FusionGhost-Gait reaches up to 67.91\% Rank-1 while remaining compact (108,970--133,674 parameters and 0.035--0.052 billion floating-point operations (GFLOPs)), achieving more than 65\% fewer parameters and 75\% lower FLOPs than existing baselines.
These results indicate that the proposed model generalizes effectively to
challenging Gulf-clothing conditions while remaining lightweight and
deployable for surveillance and access-control biometric systems.
\end{abstract}

\begin{keyword}
Gait recognition \sep Skeleton-based biometrics \sep Spatiotemporal graph convolution \sep Efficient graph neural networks \sep Clothing variation
\end{keyword}

\end{frontmatter}

\section{Introduction}
\label{sec:introduction}

Gait recognition is attractive for surveillance and access-control settings because
it can identify individuals at a distance without subject cooperation, even under low resolution,
viewpoint variation, and long-range capture~\citep{wan2018survey,liao2020model,wang2012gei,wu2016comprehensive}.
These properties make gait suitable for border checkpoints, airport corridors, and large-scale
urban monitoring where face or fingerprint acquisition is often impractical.

A central challenge is \emph{clothing variation}, which refers to gait
appearance changes caused by different garments and carried items that
alter visible body shape and motion cues. However, not all clothing
covariates are equivalent. Standard gait benchmarks mainly emphasize
coats, jackets, and bags~\citep{yu2006framework,takemura2018multi}, while
in Gulf clothing contexts the \emph{Kandoora} and \emph{Abaya} create a
more severe occlusion regime in which flowing ankle-length garments hide
lower-limb articulation and degrade the pose signal itself rather than
merely altering appearance.

This distinction matters especially for skeleton-based methods. Skeleton-based
recognition reduces appearance bias by modeling motion as joint
trajectories rather than raw pixels~\citep{yan2018spatial,liao2020model,teepe2021gaitgraph},
however, this benefit depends on reliable pose extraction. Under Gulf
clothing, the recognition problem shifts upstream. Lower-body keypoints
become noisy, unstable, or missing, and those errors propagate directly
into graph-based gait representations.

To study this underrepresented regime, we propose \emph{ArabGait}, a
45-subject indoor multi-view dataset with four walking conditions and a
targeted cross-clothing protocol focused on Gulf-clothing occlusion.

Methodologically, the backbone line evolved from ST-GCN~\citep{yan2018spatial}
to ResGCN~\citep{song2020stronger}, then to GaitGraph/GaitGraph2 for gait
recognition~\citep{teepe2021gaitgraph,teepe2022towards}. In this
pose-noisy regime, higher-capacity backbones can overfit and fail to
generalize across clothing conditions, and multi-stream fusion can
amplify noise from derived bone and velocity streams instead of improving
robustness.

These observations motivate a more capacity-aware architecture. In
pose-noisy gait settings, repeated graph and temporal processing must be
modeled compactly to avoid unnecessary channel expansion and unstable
temporal fitting. We therefore introduce \emph{FusionGhost-Gait}, which
models graph interaction and lightweight temporal dependency inside one
fused residual unit, uses compact temporal Ghost kernels to reduce
redundant channel mixing and sensitivity to lower-limb noise, and
preserves efficient temporal downsampling through a lightweight stride
stage.

Our evaluation strategy has two complementary roles. Public datasets are
used for standard comparability to prior work, while ArabGait is used for
targeted robustness analysis under severe Gulf-clothing occlusion.
ArabGait is therefore not intended to replace standard public cross-view
benchmarks, but to isolate a clothing regime that they underrepresent.

Using this benchmark setting, we study capacity-sensitive robustness
under Gulf-clothing occlusion and design \emph{FusionGhost-Gait} for
pose-noisy skeleton trajectories.
In this paper, we make the following contributions.
\begin{enumerate}
\item \textbf{Problem formulation and benchmark.} We introduce ArabGait, a 45-subject multi-view dataset and cross-clothing evaluation protocol that isolates severe Gulf-clothing-induced pose degradation, and we quantify how lower-limb pose reliability deteriorates under this regime.
\item \textbf{FusionGhost-Gait method and architecture.} We introduce \emph{FusionGhost-Gait}, a fused spatial--temporal Ghost design that models graph interaction and lightweight temporal dependency within a single residual unit using compact temporal Ghost kernels.
\item \textbf{Empirical insight and validation.} We combine CASIA-B cross-view comparison with ArabGait experiments to show that pose degradation and multi-stream noise amplification are central failure modes, and that the fused design provides a strong accuracy--efficiency operating point under cross-clothing evaluation.
\end{enumerate}

\section{Related Work}
\label{sec:related}

\subsection{Cross-Covariate Gait Recognition}

Clothing and carrying conditions remain central challenges in gait recognition.
Benchmarks such as CASIA-B~\citep{yu2006framework} introduced normal, bag, and coat
conditions, while larger datasets including OU-MVLP~\citep{takemura2018multi} and GREW~\citep{guo2025gaitbase}
enabled large-scale view- and appearance-invariant modeling.
Silhouette-based approaches such as GaitSet~\citep{chao2019gaitset},
GaitPart~\citep{fan2020gaitpart}, and GEINet~\citep{shiraga2016geinet}
learn set-level or part-level representations to mitigate appearance variation.
Extensions incorporating temporal aggregation or attention include
CSTL~\citep{cstl2021}, GaitSlice~\citep{gaitslice2022}, AttenGait~\citep{castro2024attengait},
and HGRBOL~\citep{hgrbol2023}. Despite strong performance, silhouette-based methods
remain tied to pixel-level appearance, which becomes unreliable under heavy-occlusion clothing
such as Gulf clothing.

\subsection{Skeleton-Based and GCN Gait Recognition}

Graph convolutional networks (GCNs) dominate skeleton-based modeling.
ST-GCN~\citep{yan2018spatial} introduced joint spatial--temporal graph convolutions,
followed by multi-stream and topology-refinement architectures~\citep{shi2019two,song2020stronger,cheng2020shiftgcn,chen2021channel}.
GaitGraph~\citep{teepe2021gaitgraph} and GaitGraph2~\citep{teepe2022towards}
adapted ResGCN~\citep{song2020stronger} for gait recognition using contrastive learning.
Additional directions explore biomechanical or physics-aware representations
and pose normalization strategies~\citep{rashmi2022gait,gaitsg2023,guo2023paa,gaittr2023}.
Lightweight skeleton models and attention-enhanced graph networks aim to balance
efficiency and accuracy~\citep{agsgcn2022}.
While skeleton-based methods improve privacy and robustness to appearance,
they implicitly rely on stable pose estimation, which degrades under Gulf clothing.

\subsection{Efficient Graph and Temporal Modeling}

Efficiency has become critical for deployment on edge and real-time systems.
GhostNet~\citep{han2020ghostnet} demonstrated that redundant feature maps can be replaced
by inexpensive linear operations, inspiring lightweight design strategies.
In skeleton modeling, approaches such as Shift-GCN~\citep{cheng2020shiftgcn}
and channel-adaptive topology refinement~\citep{chen2021channel}
reduce computational cost by modifying spatial reasoning.
However, existing work largely overlooks redundancy within the temporal--spatial
bottleneck structure of ResGCN blocks, particularly the interaction between SCN
and TCN components. We therefore target structured efficiency under occlusion,
using Ghost-based temporal representations inside a fused spatiotemporal block design.
Existing gait benchmarks emphasize standard clothing and view variation,
while existing ResGCN-style skeleton models are not designed specifically
for severe clothing-induced pose degradation. This motivates both a
targeted benchmark and a capacity-controlled architectural redesign.
\section{ArabGait and the Cross-Clothing Evaluation Setting}
\label{sec:dataset}

\subsection{Data Collection Protocol}
We collected a gait dataset of 45 subjects in a controlled indoor environment.
This study introduces a new dataset specifically for evaluating gait recognition under Gulf-clothing occlusion.
This study was approved by the American University of Sharjah Institutional Review Board under Protocol \#24-039, and written informed consent was obtained from all participants before data collection.
Each subject walked along a straight 7.2-m path under four conditions.
\begin{itemize}
\item \textbf{Casual clothing:} Regular everyday clothing without accessories.
\item \textbf{Casual clothing with Bag:} Casual clothing while carrying a bag.
\item \textbf{Gulf clothing:} Wearing traditional Gulf clothing (Kandoora/Abaya),
a full-length garment that covers the body from shoulders to ankles.
\item \textbf{Gulf clothing with Bag:} Gulf clothing combined with carrying a bag.
\end{itemize}
To preserve natural gait behavior in both bag conditions, participants carried a standard laptop bag using their preferred style, either hand-held or over one shoulder.

\begin{figure}[t]
\centering
\includegraphics[width=1\linewidth]{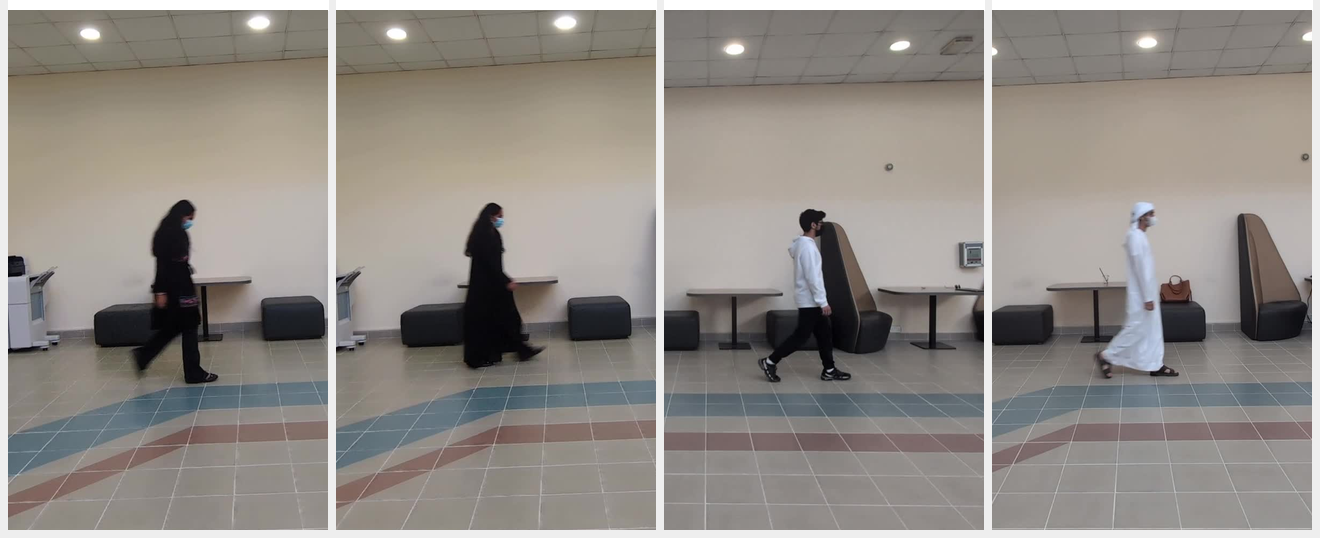}
\caption{Paired clothing examples from the dataset. The left pair shows one subject in Casual and Abaya. The right pair shows another subject in Casual and Kandoora. The Gulf-clothing samples show stronger lower-limb occlusion than Casual clothing.}
\label{fig:gulf_clothing_types}
\end{figure}

Three cameras recorded the walkway at 30 FPS from 45\textdegree,
90\textdegree, and 135\textdegree{} viewpoints. The 90\textdegree{} camera
was centered 3.8 m perpendicular to the midpoint of the path, and the two
flanking cameras were positioned 3.6 m to either side of the center camera.
Subjects walked between endpoints A and B in both directions
(A$\rightarrow$B and B$\rightarrow$A), and each subject performed multiple walks per condition,
yielding 10 gait clips. The dataset details are provided in Table~\ref{tab:dataset_stats}.
Fig.~\ref{fig:gulf_clothing_types} presents two paired examples under matched capture settings.
The left pair shows one subject in Casual and Abaya, and the right pair shows another subject in
Casual and Kandoora. These pairs illustrate how Gulf clothing increases lower-limb occlusion
relative to Casual clothing.
Fig.~\ref{fig:gulf_clothing_types} also helps localize the problem more
precisely for the reader. In the Casual samples, the leg contour and the
relative separation between the knees, shins, and ankles remain visually
available, so the walking pattern is easier to recover from pose
estimation. In the Abaya and Kandoora samples, that structure is partially
hidden by the flowing garment, especially around the knees and ankles where
gait cues are most informative. The visual issue is therefore not only a
change in overall appearance. It is a direct loss of observable lower-limb
articulation, which explains why pose extraction becomes less stable under
Gulf clothing.

\begin{table}[!t]
\caption{Dataset Statistics}
\label{tab:dataset_stats}
\centering
\footnotesize
\setlength{\tabcolsep}{4pt}
\resizebox{\columnwidth}{!}{%
\begin{tabular}{@{}l l@{}}
\toprule
\textbf{Property} & \textbf{Value} \\
\midrule
Number of subjects & 45 \\
Age range & 18--25 years \\
Gender distribution & 27 men and 18 women \\
\multirow{2}{*}{Walking conditions} & 4 (Casual clothing, Casual clothing with Bag,\\
 & Gulf clothing, Gulf clothing with Bag) \\
Camera views & 3 (45\textdegree, 90\textdegree, 135\textdegree) \\
Walks per condition & 10 walks (5 A$\rightarrow$B, 5 B$\rightarrow$A) \\
Train/test split & Subject-disjoint \\
Skeleton format & COCO 17-keypoint \\
Sequence length & 60 frames \\
\bottomrule
\end{tabular}
}
\end{table}

\subsection{ArabGait as a Benchmark for Gulf-Clothing Occlusion}
ArabGait is not intended as a universal gait benchmark. Its purpose is to
isolate a clothing regime that standard datasets underrepresent, namely severe
lower-limb occlusion caused by Gulf garments. This focus is motivated by
the practical visual conditions of the Gulf region, where garments such as
the Kandoora and Abaya are common in everyday public spaces. In settings
such as university campuses, airports, transport hubs, government
buildings, and large pedestrian walkways, gait systems are more likely to
encounter full-length flowing garments than the coat-style covariates that
dominate standard public benchmarks. As a result, a benchmark that reflects
this clothing regime is important for evaluating whether skeleton-based gait
models remain reliable in the environments where they may actually be used.
The challenge lies not only in the occlusion of lower-limb motion cues by
these garments, but also in the substantial appearance shift when the same
person transitions from Casual clothing to an Abaya or a Kandoora. That
shift makes the cross-clothing setting more demanding even before the
downstream effects on pose estimation are considered.

This distinction matters because Gulf clothing does not simply change
appearance at the image level. It often alters the quality of the extracted
pose itself by obscuring the knees, ankles, and lower-leg motion that are
central to gait dynamics. In other words, the challenge is not only
cross-appearance recognition, but recognition under systematic degradation
of the structural motion signal. That setting is materially different from
the jacket, coat, and bag variations emphasized in datasets such as
CASIA-B, where enough lower-limb visibility is usually preserved for more
stable keypoint extraction.

The value of ArabGait therefore lies not only in its four clothing
conditions and three views, but in its ability to isolate a targeted and
regionally relevant failure mode. Rather than serving as a replacement for
established public benchmarks, ArabGait complements them by enabling a
focused study of how Gulf-region clothing affects pose reliability and how
that degradation propagates into skeleton-based recognition under explicit
cross-clothing transfer.

\subsection{Pose Extraction and Reliability Analysis}
We extract 2D pose sequences using HRNet-W48~\citep{sun2019deep}, a high-resolution network
that maintains multi-scale representations throughout. For each video frame, HRNet outputs
17 COCO-format keypoints with associated confidence scores.

Pose estimation quality varies significantly across conditions. Under Casual clothing and Casual clothing with Bag conditions,
HRNet produces stable, high-confidence keypoints for all joints. Under Gulf clothing,
lower-limb joints (hips, knees, ankles) exhibit substantially reduced confidence,
increased positional drift between frames, and occasional complete failures. This
degradation is most pronounced for the ankle and knee joints, which are fully
occluded by the Kandoora and Abaya.

For pose-quality analysis, we use a confidence threshold of 0.40 to mark
low-confidence detections as unreliable. This threshold is used to quantify
pose degradation (e.g., confidence and failure-rate statistics).

\begin{table}[!t]
\caption{Pose-quality summary across all views (HRNet confidences).}
\label{tab:pose_quality_metrics}
\centering
\scriptsize
\setlength{\tabcolsep}{3pt}
\resizebox{\columnwidth}{!}{%
\begin{tabular}{lcccc}
\toprule
\textbf{Condition} & \textbf{Upper Conf.} & \textbf{Lower Conf.} & \textbf{Lower Fail (\%)} & \textbf{Lower Std.} \\
\midrule
Casual clothing & 0.831 & 0.811 & 7.98 & 0.046 \\
Casual clothing with Bag & 0.801 & 0.766 & 11.83 & 0.056 \\
Gulf clothing & 0.780 & 0.729 & 13.03 & 0.059 \\
Gulf clothing with Bag & 0.740 & 0.683 & 17.82 & 0.064 \\
\bottomrule
\end{tabular}
}
\end{table}

\begin{figure}[!t]
\centering
\includegraphics[width=0.78\columnwidth,clip]{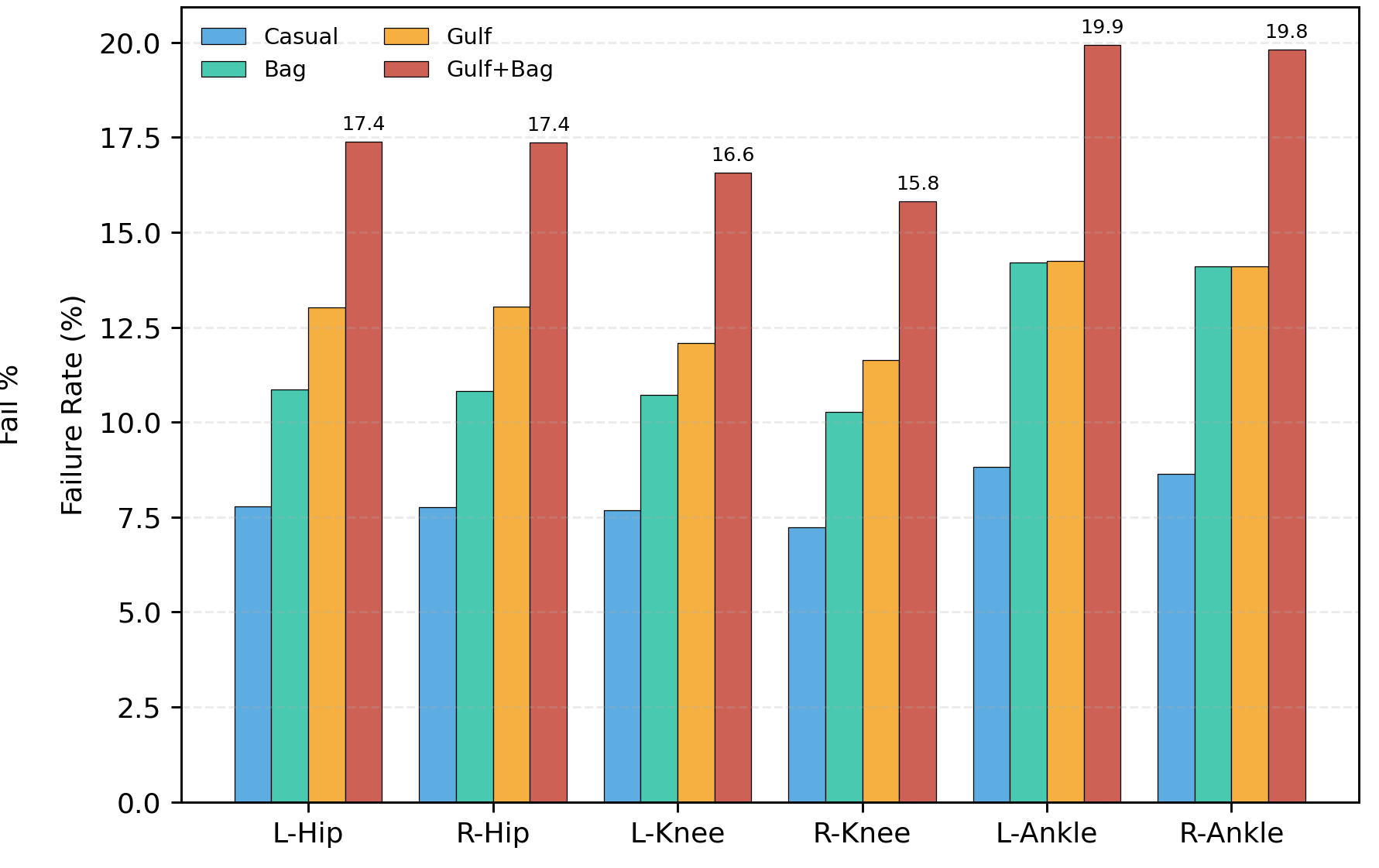}
\caption{Per-joint lower-limb failure profile across all four conditions from the full HRNet output set.}
\label{fig:pose_quality_overview}
\end{figure}

Table~\ref{tab:pose_quality_metrics} and Fig.~\ref{fig:pose_quality_overview} indicate that performance loss
is concentrated in lower-limb joints and intensifies under Gulf clothing
occlusion. Across all views, lower-limb mean confidence drops from 0.811 (Casual clothing)
to 0.683 (Gulf clothing with Bag), while lower-limb failure rate rises from 7.98\% to 17.82\%.
Fig.~\ref{fig:pose_quality_overview} shows consistent escalation from Casual clothing to Gulf clothing with Bag, with
ankles exhibiting the highest failure rates.

Because the dominant corruption appears at pose extraction time, the
methodological goal is not only view or appearance invariance, but also
robustness to pose-noisy and partially unreliable joint trajectories.

\section{Proposed Method: FusionGhost-Gait}
\label{sec:method}

\subsection{Design Motivation}
\label{sec:baseline_variants}

Severe Gulf-clothing occlusion corrupts lower-limb pose trajectories and
destabilizes the derived structural cues used by skeleton-based gait
models. The architectural goal is therefore not only efficient
spatiotemporal modeling, but robustness to pose-noisy and partially
unreliable joint sequences under constrained model size.

FusionGhost-Gait addresses this setting with a fused spatio\-temporal skeleton
backbone built from repeated Ghost-based residual blocks. The two reported
configurations in this paper are FusionGhost-Gait-Nano and FusionGhost-Gait-S.
We report these two configurations to bracket the capacity-sensitive operating
region in this pose-noisy regime: FusionGhost-Gait-Nano (3 fused bottlenecks)
as the compact endpoint and FusionGhost-Gait-S (6 fused bottlenecks) as the
moderate-capacity endpoint. Our analysis indicates that under severe
Gulf\-clothing occlusion, moving beyond moderate depth does not consistently
improve cross-clothing generalization and may increase sensitivity to noisy
lower\-limb trajectories. Reporting Nano and S captures the practical
accuracy\--efficiency operating range with one shared fused block design while
avoiding deeper variants that increase compute without stable robustness gains.

\subsection{FusionGhost-Gait Overview}
\label{sec:ghost_variants}

FusionGhost-Gait is organized into four backbone stages, each composed of
repeated fused spatiotemporal Ghost blocks. In the single-stream setting,
the backbone operates on joint trajectories only. The Spatiotemporal Ghost
Stem (STG-S) serves as the
input stem of each stream. It performs an initial Ghost down projection
followed by graph interaction to lift raw joint trajectories into the
backbone feature space. Unlike the Spatiotemporal Ghost Bottleneck
(STG-B), STG-S is used for early feature
extraction and does not include the full fused bottleneck structure used
in the deeper stages. In the multi-stream setting, we use a standard
three-stream skeleton input composed of joints, bones, and motion. Each
stream is processed independently through STG-S and the first two STG-B
blocks. Fusion is applied after the second STG-B block, and the fused
representation is then processed by the remaining shared backbone
stages. Both reported configurations (Nano and S) preserve one fused
block design and differ only in bottleneck depth.

\subsection{Fused Spatiotemporal Ghost Block}

\begin{figure}[!t]
\centering
\resizebox{0.70\textwidth}{!}{\input{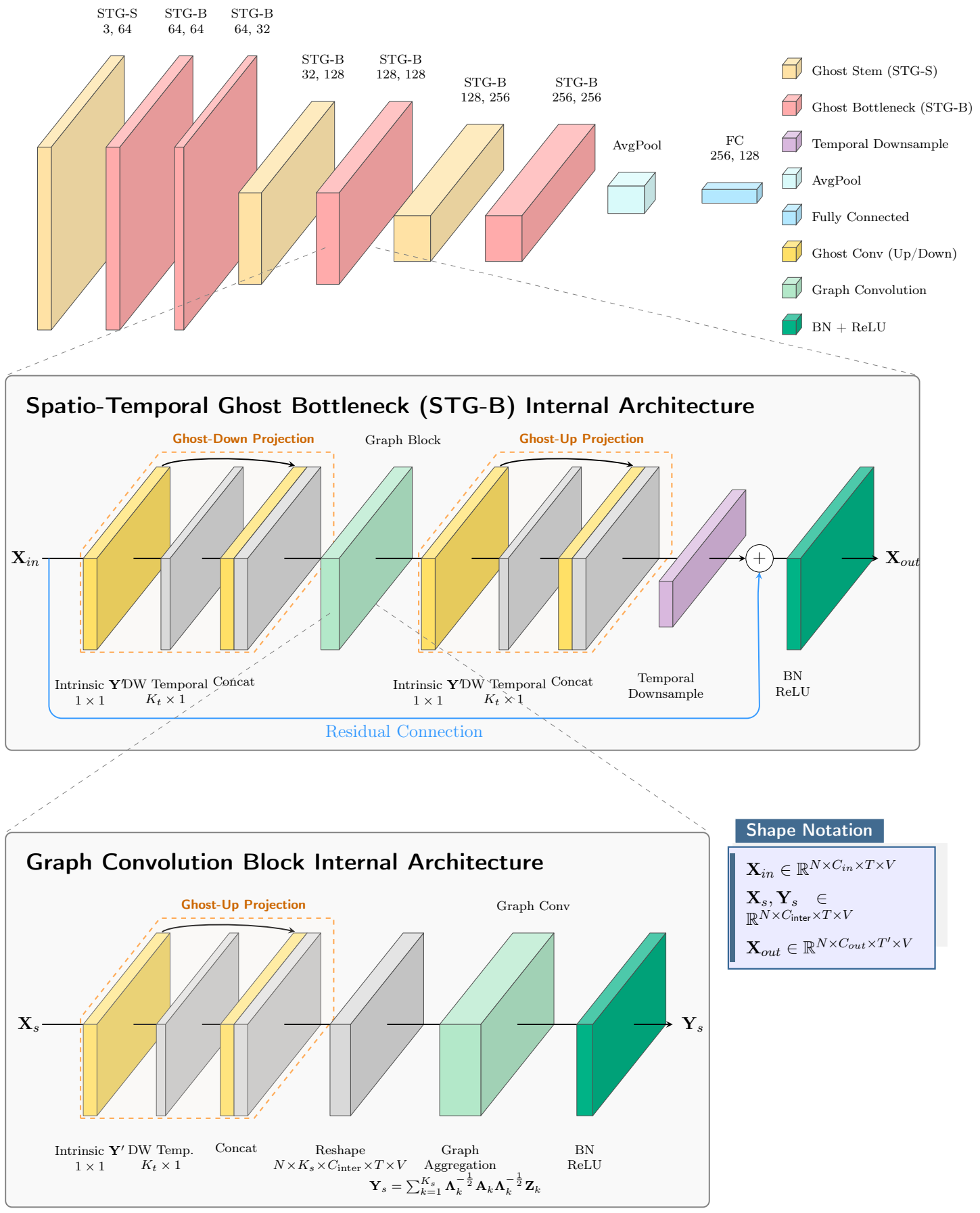}}
\caption{FusionGhost-Gait-S architecture diagram. The backbone is composed of a Ghost stem (STG-S), repeated spatiotemporal Ghost bottlenecks (STG-B), lightweight temporal downsampling stages, and a final pooling/embedding head. Both reported FusionGhost-Gait configurations (Nano and S) share the fused STG-B design shown in the lower panel and differ only in depth. The figure shows the shared FusionGhost-Gait-S backbone, while the multi-stream fusion scheme is described in the text.}
\label{fig:fusionghost_gait_s_arch}
\end{figure}

Fig.~\ref{fig:fusionghost_gait_s_arch} shows the staged FusionGhost-Gait-S backbone and
the proposed spatiotemporal Ghost bottleneck (STG-B), where temporal
dependency is modeled inside the block through compact Ghost kernels.

\subsection{Formal Definitions}
\label{sec:formal_defs}

To match Fig.~\ref{fig:fusionghost_gait_s_arch}, we use the same tensor names shown in the
diagram. The STG-B input is
$\mathbf{X}_{in}\in\mathbb{R}^{N\times C_{in}\times T\times V}$ and the block output is
$\mathbf{X}_{out}\in\mathbb{R}^{N\times C_{out}\times T'\times V}$, where $N$ is batch
size, $T$ is the number of frames, and $V$ is the number of joints.
Let $s_t$ denote the temporal stride of the block. Then $T'=T$ when
$s_t=1$, and otherwise $T'$ is the temporally downsampled length produced
by stride $s_t$.
Channel projections are applied independently at each $(t,v)$ location, while
adjacency multiplication acts on the joint axis $V$. We omit bias terms and
use $\sigma(\cdot)$ for BN+activation.

\subsubsection{Ghost Projections}
Both the channel-reduction projection before the graph block and the
channel-expansion projection after it are Ghost projections with Ghost
ratio $r=2$. We set $r=2$ to balance representational capacity and
efficiency, half
of the channels are kept as intrinsic channels for stable graph features,
while the other half are generated as Ghost channels for low-cost
temporal enrichment. For a target width $C_q$ with $q\in\{d,u\}$ denoting
the down and up projections, we split the output channels as
\begin{equation}
\label{eq:fsg_channels}
C_q=C_q^{p}+C_q^{g},
\qquad
C_q^{p}=\frac{C_q}{r},
\qquad
C_q^{g}=C_q-C_q^{p},
\end{equation}
where $C_q^{p}$ is the number of intrinsic channels produced by a standard
pointwise projection and $C_q^{g}$ is the number of Ghost channels produced
by a cheap temporal operator. Since $r=2$ in this work, the intrinsic and
Ghost parts each contribute half of the target width.

We define a Ghost projection operator $\Phi_G(\cdot)$ as
\begin{align}
\label{eq:fsg_proj}
\mathbf{Z}_{q}^{(p)} &= \sigma(\mathbf{X}\mathbf{W}_{q}^{(p)}),
& \mathbf{W}_{q}^{(p)} &\in \mathbb{R}^{C_{src}\times C_q^{p}}, \notag\\
\mathbf{Z}_{q}^{(g)} &= \mathcal{K}_{t}^{(q)}\!\left(\mathbf{Z}_{q}^{(p)}\right),
& \mathbf{Z}_{q}^{(g)} &\in \mathbb{R}^{N\times C_q^{g}\times T_q\times V}, \notag\\
\Phi_G(\mathbf{X};q) &= [\mathbf{Z}_{q}^{(p)}\,\|\,\mathbf{Z}_{q}^{(g)}],
& \Phi_G(\mathbf{X};q) &\in \mathbb{R}^{N\times C_q\times T_q\times V}.
\end{align}
Here, $\mathbf{Z}_{q}^{(p)}$ denotes the primary features,
$\mathbf{Z}_{q}^{(g)}$ denotes the Ghost features,
$\mathbf{W}_{q}^{(p)}$ is the primary pointwise projection,
$\mathcal{K}_{t}^{(q)}$ is a depthwise temporal convolution with kernel size
$(k_t,1)$, and $C_{src}$ is the source-channel width of the projection. The
temporal kernel acts only on the primary features and generates the Ghost
features. It does not act on the entire tensor before the pointwise mapping.
For the Ghost-down projection, $C_{src}=C_{in}$ and $T_q=T$. For the
Ghost-up projection, $C_{src}=C_g$ and $T_q=T$ before any temporal stride is
applied.

\subsubsection{Graph Convolution Unit}
Spatial interaction is modeled with partitioned adjacency aggregation,
following the ST-GCN formulation~\citep{yan2018spatial}. For each frame,
the same normalized adjacency partitions are used:
\begin{equation}
\label{eq:fsg_graph}
\mathbf{X}_{d}=\Phi_G(\mathbf{X}_{in};d),
\qquad
\mathbf{Y}_{g} = \sum_{k=1}^{K}\hat{\mathbf{A}}_{k}\,\mathbf{X}_{d}\,\mathbf{W}_{k},
\end{equation}
where $\hat{\mathbf{A}}_{k}$ is the normalized adjacency of partition $k$ and
$\mathbf{W}_{k}\in\mathbb{R}^{C_d\times C_g}$ is the learnable channel
transform. Here, $\mathbf{X}_{d}\in\mathbb{R}^{N\times C_d\times T\times V}$ is the
Ghost-down output with $C_d=C_{inter}$, and
$\mathbf{Y}_{g}\in\mathbb{R}^{N\times C_g\times T\times V}$ is the output of the
graph interaction stage.
This corresponds to the Graph Block in Fig.~\ref{fig:fusionghost_gait_s_arch}.

\subsubsection{Ghost-Up Projection}
The block then forms its output representation through a Ghost-up
projection:
\begin{equation}
\label{eq:fsg_up}
\mathbf{X}_{f}=\Phi_G(\mathbf{Y}_{g};u),
\qquad
\mathbf{X}_{f}\in\mathbb{R}^{N\times C_u\times T\times V},
\end{equation}
where $C_u=C_{out}$ for the up projection. Thus, the graph interaction output
is first mapped to intrinsic channels by $\mathbf{W}_{u}^{(p)}$ and then
augmented with Ghost channels generated by the temporal kernel
$\mathcal{K}_{t}^{(u)}$.

\subsubsection{Fused Spatiotemporal Ghost Block}
The full STG-B is therefore a Ghost-down projection, followed by a
graph interaction stage, followed by a Ghost-up projection, and then the
temporal stride and residual addition:
\begingroup
\footnotesize
\begin{align}
\label{eq:fsg_block}
\mathcal{D}_{t}(\mathbf{Z}) &=
\begin{cases}
\mathbf{Z}, & s_t=1,\\
\mathrm{Down}_{t}(\mathbf{Z}; s_t), & s_t>1,
\end{cases}
\notag\\
\mathcal{R}(\mathbf{X}_{in}) &=
\begin{cases}
\mathbf{X}_{in}, & C_{in}=C_{out}\ \text{and}\ s_t=1,\\
\mathrm{Proj}_{skip}(\mathbf{X}_{in}; C_{out}, s_t), & \text{otherwise},
\end{cases}
\notag\\
\mathbf{X}_{out} &= \mathcal{D}_{t}(\mathbf{X}_{f}) + \mathcal{R}(\mathbf{X}_{in}). \notag
\end{align}
\endgroup
Here, $\mathbf{X}_{f}\in\mathbb{R}^{N\times C_{out}\times T\times V}$ is the Ghost-up
output before temporal downsampling. The operator $\mathrm{Down}_{t}(\cdot; s_t)$ denotes the
lightweight temporal stride stage used only when $s_t>1$, while
$\mathrm{Proj}_{skip}(\cdot; C_{out}, s_t)$ denotes the residual projection used
when the block changes temporal resolution or channel width.
The temporal Ghost operator injects lightweight temporal context in both
the Ghost-down and Ghost-up projections, while the graph block performs the
joint-wise aggregation within the fused block.

\section{Experimental Setup}
\label{sec:setup}
\subsection{Public Benchmark Evaluation}
\label{sec:public_protocol}

To provide standard comparability to prior work, we report CASIA-B
cross-view results using the conventional NM/BG/CL evaluation split. We
follow the original CASIA-B protocol, reproduce prior-method rows from
the cited literature, and report FusionGhost-Gait rows from our own
runs under the same standard evaluation.
We use these public-benchmark results only as external grounding for the
fused design; the targeted clothing-robustness analysis is conducted on
ArabGait.

\subsection{ArabGait Cross-Clothing Evaluation}
\label{sec:protocol}

ArabGait uses a subject-disjoint split with IDs 01--35 for
training/validation and IDs 36--45 for testing. We retain the original
same-view protocol as a legacy reference where, for each test subject and view,
the gallery uses Casual sequences \#1--6, while probes use Casual
sequences \#7--10 plus all Casual with Bag, Gulf, and Gulf with Bag
sequences.

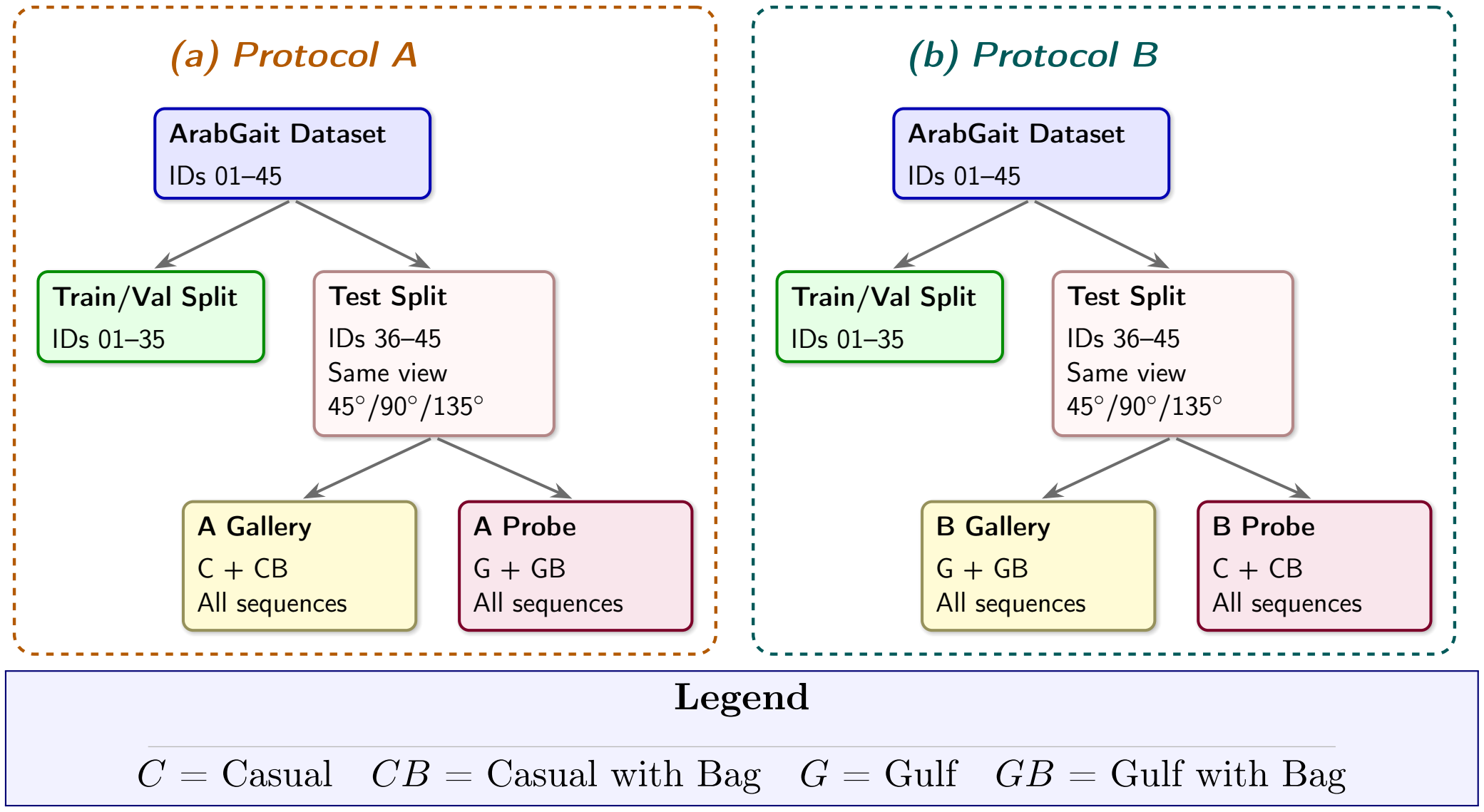
\begin{figure}[!t]
\centering
\begin{minipage}[t]{0.49\columnwidth}
\centering
\resizebox{\textwidth}{!}{%
\def\nodesepline{\\[1.2mm]}
\begin{tikzpicture}[
  font=\sffamily\small,
  flow/.style={-{Stealth[length=3.0mm, width=2.2mm]}, line width=1.15pt, draw=gray!85!black, shorten <=1.5pt, shorten >=1.5pt},
  box/.style={
    rounded corners=4pt,
    line width=1.15pt,
    align=left,
    inner sep=5.5pt,
    blur shadow={shadow blur steps=5, shadow xshift=0.7pt, shadow yshift=-0.7pt, shadow blur radius=2pt}
  },
  panel/.style={
    rounded corners=8pt,
    line width=1.25pt,
    dashed,
    draw=orange!70!black,
    fill=none,
    inner sep=9pt
  },
  title/.style={font=\bfseries\itshape\sffamily\large, text=orange!70!black},
  root/.style={box, fill=blue!10, draw=blue!70!black, text width=3.5cm},
  train/.style={box, fill=green!10, draw=green!55!black, text width=2.8cm},
  test/.style={box, fill=pink!12, draw=pink!70!black, text width=3.0cm},
  gallery/.style={box, fill=yellow!20, draw=yellow!50!black, text width=2.9cm},
  probe/.style={box, fill=purple!10, draw=purple!65!black, text width=2.9cm}
]
\node[title] (titleA) at (0,0) {(a)\;Protocol A};
\node[root] (dataA) at (0,-1.35) {
  \textbf{ArabGait Dataset}\nodesepline
  IDs 01--45
};
\node[train, anchor=north] (trainA) at (-2.0,-3.0) {
  \textbf{Train/Val Split}\nodesepline
  IDs 01--35
};
\node[test, anchor=north] (testA) at (2.0,-3.0) {
  \textbf{Test Split}\nodesepline
  IDs 36--45\\
  Same view\\
  45$^\circ$/90$^\circ$/135$^\circ$
};
\node[gallery, anchor=north] (gA) at (0.1,-6.25) {
  \textbf{A Gallery}\nodesepline
  C + CB\\
  All sequences
};
\node[probe, anchor=north] (pA) at (4.0,-6.25) {
  \textbf{A Probe}\nodesepline
  G + GB\\
  All sequences
};
\draw[flow] (dataA.south) -- (trainA.north);
\draw[flow] (dataA.south) -- (testA.north);
\draw[flow] (testA.south) -- (gA.north);
\draw[flow] (testA.south) -- (pA.north);
\node[panel, fit=(titleA)(dataA)(trainA)(testA)(gA)(pA)] {};
\end{tikzpicture}
}
\end{minipage}
\hfill
\begin{minipage}[t]{0.49\columnwidth}
\centering
\resizebox{\textwidth}{!}{%
\def\nodesepline{\\[1.2mm]}
\begin{tikzpicture}[
  font=\sffamily\small,
  flow/.style={-{Stealth[length=3.0mm, width=2.2mm]}, line width=1.15pt, draw=gray!85!black, shorten <=1.5pt, shorten >=1.5pt},
  box/.style={
    rounded corners=4pt,
    line width=1.15pt,
    align=left,
    inner sep=5.5pt,
    blur shadow={shadow blur steps=5, shadow xshift=0.7pt, shadow yshift=-0.7pt, shadow blur radius=2pt}
  },
  panel/.style={
    rounded corners=8pt,
    line width=1.25pt,
    dashed,
    draw=teal!70!black,
    fill=none,
    inner sep=9pt
  },
  title/.style={font=\bfseries\itshape\sffamily\large, text=teal!70!black},
  root/.style={box, fill=blue!10, draw=blue!70!black, text width=3.5cm},
  train/.style={box, fill=green!10, draw=green!55!black, text width=2.8cm},
  test/.style={box, fill=pink!12, draw=pink!70!black, text width=3.0cm},
  gallery/.style={box, fill=yellow!20, draw=yellow!50!black, text width=2.9cm},
  probe/.style={box, fill=purple!10, draw=purple!65!black, text width=2.9cm}
]
\node[title] (titleB) at (0,0) {(b)\;Protocol B};
\node[root] (dataB) at (0,-1.35) {
  \textbf{ArabGait Dataset}\nodesepline
  IDs 01--45
};
\node[train, anchor=north] (trainB) at (-2.0,-3.0) {
  \textbf{Train/Val Split}\nodesepline
  IDs 01--35
};
\node[test, anchor=north] (testB) at (2.0,-3.0) {
  \textbf{Test Split}\nodesepline
  IDs 36--45\\
  Same view\\
  45$^\circ$/90$^\circ$/135$^\circ$
};
\node[gallery, anchor=north] (gB) at (0.1,-6.25) {
  \textbf{B Gallery}\nodesepline
  G + GB\\
  All sequences
};
\node[probe, anchor=north] (pB) at (4.0,-6.25) {
  \textbf{B Probe}\nodesepline
  C + CB\\
  All sequences
};
\draw[flow] (dataB.south) -- (trainB.north);
\draw[flow] (dataB.south) -- (testB.north);
\draw[flow] (testB.south) -- (gB.north);
\draw[flow] (testB.south) -- (pB.north);
\node[panel, fit=(titleB)(dataB)(trainB)(testB)(gB)(pB)] {};
\end{tikzpicture}
}
\end{minipage}
\vspace{1mm}

\begin{minipage}{\columnwidth}
\centering
\fcolorbox{blue!45!black}{blue!5}{%
\begin{minipage}{0.98\columnwidth}
\centering
\footnotesize
\textbf{Legend}\\[-0.7mm]
\textcolor{gray!55}{\rule{0.82\linewidth}{0.25pt}}\\[-0.6mm]
$C$ = Casual \hspace{1.2mm}
$CB$ = Casual with Bag \hspace{1.2mm}
$G$ = Gulf \hspace{1.2mm}
$GB$ = Gulf with Bag
\end{minipage}}
\end{minipage}
\caption{ArabGait cross-clothing evaluation protocols under identical-view matching. IDs 01--35 are used for training and validation, and IDs 36--45 are used for testing. Protocol A uses C+CB as gallery and G+GB as probe. Protocol B swaps gallery and probe.}
\label{fig:cross_clothing_split}
\end{figure}

Our main ArabGait evaluation uses two
cross-clothing transfer settings under identical-view matching. Protocol
A uses Casual plus Casual-with-Bag sequences as gallery and Gulf plus
Gulf-with-Bag sequences as probe. Protocol B swaps the roles, using Gulf
plus Gulf-with-Bag as gallery and Casual plus Casual-with-Bag as probe.
The training split is unchanged; only the ArabGait test protocol changes
for cross-clothing evaluation. These two protocols are shown in
Fig.~\ref{fig:cross_clothing_split}. Same-view matching is used here
because ArabGait is intended to isolate clothing-induced pose degradation
rather than to serve as a full cross-view benchmark.

For each condition and view angle
\(\theta \in \{45^\circ, 90^\circ, 135^\circ\}\), we compute
nearest-neighbor Rank-1 identification with Euclidean distance. Because
each condition contributes matched sequences across the three views, we
report means across views and paired conditions:
\begin{align}
\mathrm{Acc}_{c} &= \frac{1}{|\Theta|}\sum_{\theta\in\Theta}\mathrm{Acc}_{c,\theta}, \\
\mathrm{Overall}_{A} &= \frac{\mathrm{Acc}_{G} + \mathrm{Acc}_{GB}}{2}, \\
\mathrm{Overall}_{B} &= \frac{\mathrm{Acc}_{C} + \mathrm{Acc}_{CB}}{2}.
\end{align}
Here, $\Theta=\{45^\circ,90^\circ,135^\circ\}$.

\subsection{Training Details}
\label{sec:training_details}

All models are trained using PyTorch Lightning with the training hyperparameter configuration summarized in Table~\ref{tab:hyperparams}.

\begin{table}[!t]
\caption{Training Hyperparameters}
\label{tab:hyperparams}
\centering
\footnotesize
\begin{tabular}{lc}
\toprule
\textbf{Hyperparameter} & \textbf{Value} \\
\midrule
Optimizer & AdamW \\
Learning rate & 0.005 \\
Weight decay & $1\times10^{-5}$ \\
Training loss & Supervised Contrastive~\citep{khosla2020supervised} \\
Temperature ($\tau$) & 0.01 \\
Distance metric & $L_p$ (Euclidean) \\
Embedding dimension & 128 \\
Batch size & 256 \\
Epochs & 300 \\
Sequence length & 60 frames \\
Data augmentation & Temporal flip, L/R flip, joint noise \\
\bottomrule
\end{tabular}
\end{table}

Test-time augmentation (TTA) is applied via temporal flip and left-right flip, producing
three augmented embeddings that are concatenated for each input clip. All experiments are
conducted on an NVIDIA A10G GPU with 15GB VRAM.

\subsection{Metrics}

We report Rank-1 identification accuracy, parameter count, and GFLOPs.
On CASIA-B we report standard cross-view Rank-1 with NM/BG/CL
breakdowns. On ArabGait we report Protocol A and Protocol B overall scores
and their per-condition components.

\section{Results}
\label{sec:results}
\vspace{-0.6\baselineskip}
\subsection{Public Benchmark Comparison}
\label{sec:public_results}

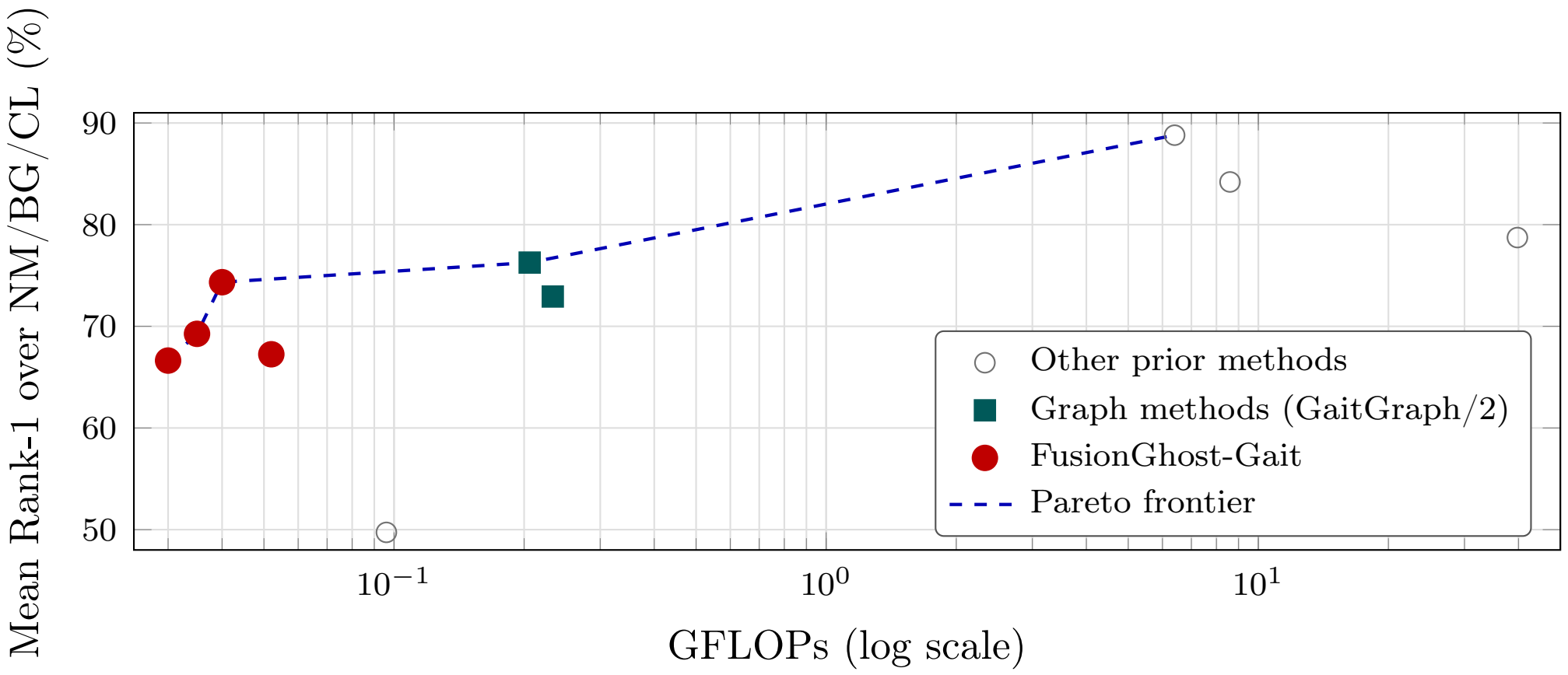
\begin{figure}[!t]
\centering
\begin{tikzpicture}
\begin{axis}[
width=\columnwidth,
height=5.3cm,
xmode=log,
xlabel={GFLOPs (log scale)},
ylabel={Mean Rank-1 over NM/BG/CL (\%)},
xmin=0.025, xmax=50,
ymin=48, ymax=91,
grid=both,
grid style={gray!25},
tick label style={font=\scriptsize},
label style={font=\footnotesize},
legend style={
font=\scriptsize,
draw=gray!65!black,
fill=white,
rounded corners=2pt,
at={(0.98,0.03)},
anchor=south east
},
legend cell align=left
]
\addplot[
only marks,
mark=o,
mark size=2.4pt,
draw=black!55,
fill=white
] coordinates {
(39.78,78.73)
(8.6,84.20)
(6.403,88.80)
(0.096,49.73)
};
\addlegendentry{Other prior methods}

\addplot[
only marks,
mark=square*,
mark size=2.5pt,
color=teal!70!black
] coordinates {
(0.233,72.93)
(0.206,76.27)
};
\addlegendentry{Graph methods (GaitGraph/2)}

\addplot[
only marks,
mark=*,
mark size=3.0pt,
color=red!75!black
] coordinates {
(0.03,66.63)
(0.04,74.34)
(0.035,69.26)
(0.052,67.25)
};
\addlegendentry{FusionGhost-Gait}

\addplot[
thick,
dashed,
color=blue!70!black,
mark=none
] coordinates {
(0.03,66.63)
(0.035,69.26)
(0.04,74.34)
(0.206,76.27)
(6.403,88.80)
};
\addlegendentry{Pareto frontier}
\end{axis}
\end{tikzpicture}
\caption{CASIA-B accuracy--complexity trade-off using mean NM/BG/CL Rank-1 versus GFLOPs. The frontier connects non-dominated points. Methods without reported FLOPs (e.g., Two-Branch) are excluded.}
\label{fig:casia_pareto}
\end{figure}

Table~\ref{tab:casia_public} provides public-benchmark grounding on
CASIA-B cross-view evaluation under the standard NM/BG/CL protocol.
FusionGhost-Gait CASIA-B rows are trained and evaluated under the same
standard cross-view protocol used in prior GaitGraph-style CASIA-B
comparisons, while prior-method rows are reproduced from their cited
papers.
Appearance-based methods provide the highest absolute accuracy, but they
operate at much higher compute cost. By contrast, the proposed
FusionGhost-Gait variants target the compact operating region among
model-based methods. Relative to GaitGraph2, the
\emph{FusionGhost-Gait-Nano} multi-stream variant reduces complexity from
0.206 to 0.040 GFLOPs (\(-80.58\%\)) and from 350,822 to 110,742
parameters (\(-68.43\%\)) while maintaining competitive cross-view mean
accuracy over NM/BG/CL (\(74.34\%\) vs \(76.27\%\)). This efficiency-first
positioning is visualized in Fig.~\ref{fig:casia_pareto}, where
FusionGhost-Gait occupies a strong low-FLOPs region rather than pursuing
absolute top accuracy at high compute. We use these public results as
external grounding so the architectural claim is not supported only by
ArabGait.

\begin{table}[!t]
\caption{CASIA-B representative prior-method comparison under the standard NM/BG/CL protocol. Prior-method rows are reproduced from the cited literature, and FusionGhost-Gait rows are our own runs.}
\label{tab:casia_public}
\centering
\footnotesize
\setlength{\tabcolsep}{3pt}
\resizebox{\columnwidth}{!}{%
\begin{tabular}{@{}llccc@{}}
\toprule
\textbf{Type} & \textbf{Method} & \textbf{NM (\%)} & \textbf{BG (\%)} & \textbf{CL (\%)} \\
\midrule
\multirow{3}{*}{Appearance-based} & GaitNet~\citep{song2019gaitnet} & 91.6 & 85.7 & 58.9 \\
 & GaitSet~\citep{chao2019gaitset} & 95.0 & 87.2 & 70.4 \\
 & GaitPart~\citep{fan2020gaitpart} & \textbf{96.2} & \textbf{91.5} & 78.7 \\
\midrule
\multirow{7}{*}{Model-based} & PoseGait~\citep{liao2020posegait} & 68.7 & 44.5 & 36.0 \\
 & GaitGraph2~\citep{teepe2022towards} & 87.7 & 74.8 & 66.3 \\
 & GaitGraph~\citep{teepe2021gaitgraph} & 82.0 & 73.2 & 63.6 \\
 & FusionGhost-Gait-Nano (Single) & 79.91 & 67.62 & 52.36 \\
 & FusionGhost-Gait-Nano (Multi) & \textbf{83.30} & \textbf{73.96} & \textbf{65.77} \\
 & FusionGhost-Gait-S (Single) & 77.95 & 68.04 & 61.80 \\
 & FusionGhost-Gait-S (Multi) & 76.03 & 65.87 & 59.84 \\
\midrule
\multirow{1}{*}{Combined} & Two-Branch~\citep{wang2022twobranch} & \textbf{97.7} & \textbf{93.8} & \textbf{92.7} \\
\bottomrule
\end{tabular}
}
\end{table}

\begin{table}[!t]
\caption{Public models efficiency comparison.}
\label{tab:casia_public_efficiency}
\centering
\footnotesize
\setlength{\tabcolsep}{3pt}
\begin{tabular}{@{}lrr@{}}
\toprule
\textbf{Method} & \textbf{Params} & \textbf{GFLOPs} \\
\midrule
GaitNet~\citep{song2019gaitnet} & 16.19M & 39.78 \\
PoseGait~\citep{liao2020posegait} & 4.229M & 0.096 \\
GaitSet~\citep{chao2019gaitset} & 2.59M & 8.6 \\
GaitPart~\citep{fan2020gaitpart} & 1.204M & 6.403 \\
GaitGraph~\citep{teepe2021gaitgraph} & 425,834 & 0.233 \\
GaitGraph2~\citep{teepe2022towards} & 350,822 & 0.206 \\
FusionGhost-Gait-Nano (Single) & 90,270 & 0.03 \\
FusionGhost-Gait-Nano (Multi) & 110,742 & 0.04 \\
FusionGhost-Gait-S (Single) & 108,970 & 0.035 \\
FusionGhost-Gait-S (Multi) & 133,674 & 0.052 \\
\bottomrule
\end{tabular}
\end{table}

\subsection{ArabGait Main Results}
\label{sec:baseline_results}

\subsubsection{Single-Stream Results}

Table~\ref{tab:single_stream} shows that single-stream cross-clothing
generalization peaks at FusionGhost-Gait-Nano (71.51\% on Protocol A and
70.75\% on Protocol B). Within ResGCN baselines, the S-scale baseline
(N39) is the strongest reference (62.81\% on Protocol A and 68.19\% on
Protocol B). Across deeper baseline variants (N51, N57, N75), performance
drops on both Protocol A and Protocol B, supporting the capacity-sensitivity
claim in this pose-noisy regime.

\begin{table}[!t]
\caption{Single-stream ArabGait cross-clothing results with ResGCN baselines and FusionGhost-Gait models.}
\label{tab:single_stream}
\centering
\scriptsize
\setlength{\tabcolsep}{3pt}
\begin{tabular}{@{}l r r r @{\hspace{1.8mm}} r r r@{}}
\toprule
\textbf{Model} & \multicolumn{3}{c}{\textbf{Protocol A}} & \multicolumn{3}{c}{\textbf{Protocol B}}\\
\cmidrule(lr){2-4}\cmidrule(lr){5-7}
\textbf{Arch} & \textbf{Overall} & \textbf{G} & \textbf{GB} & \textbf{Overall} & \textbf{C} & \textbf{CB}\\
\midrule
ResGCN-N21 & 59.56 & 61.11 & 58.01 & 65.22 & 58.12 & 72.31\\
FusionGhost-Gait-Nano & 71.51 & 70.59 & 72.44 & 70.75 & 65.94 & 75.57\\
ResGCN-N39 & 62.81 & 65.36 & 60.26 & 68.19 & 63.75 & 72.64\\
ResGCN-N51 & 56.33 & 58.82 & 53.85 & 59.44 & 54.38 & 64.50\\
ResGCN-N57 & 57.15 & 59.80 & 54.49 & 56.27 & 50.00 & 62.54\\
ResGCN-N75 & 58.26 & 59.15 & 57.37 & 60.67 & 57.81 & 63.52\\
FusionGhost-Gait-S & 66.68 & 67.97 & 65.38 & 67.91 & 61.88 & 73.94\\
\bottomrule
\end{tabular}
\end{table}

\subsubsection{Multi-Stream Results}

Multi-stream fusion is even more sensitive to cross-clothing transfer.
Table~\ref{tab:multi_stream} shows that the compact Nano-scale baseline is the
strongest multi-stream reference on the cross-clothing protocols, with
62.17\% on Protocol A and 67.07\% on Protocol B, while the commonly used
S-scale reference drops to 52.28\% and 49.52\%, respectively. This behavior supports the view
that derived bone and velocity streams can amplify pose noise when the
lower-body signal is already unreliable.

\begin{table}[!t]
\caption{Multi-stream ArabGait cross-clothing results with ResGCN baselines and FusionGhost-Gait models.}
\label{tab:multi_stream}
\centering
\scriptsize
\setlength{\tabcolsep}{3pt}
\begin{tabular}{@{}l r r r @{\hspace{1.8mm}} r r r@{}}
\toprule
\textbf{Model} & \multicolumn{3}{c}{\textbf{Protocol A}} & \multicolumn{3}{c}{\textbf{Protocol B}}\\
\cmidrule(lr){2-4}\cmidrule(lr){5-7}
\textbf{Arch} & \textbf{Overall} & \textbf{G} & \textbf{GB} & \textbf{Overall} & \textbf{C} & \textbf{CB}\\
\midrule
ResGCN-N21 & 62.17 & 65.69 & 58.65 & 67.07 & 63.12 & 71.01\\
FusionGhost-Gait-Nano & 64.92 & 68.30 & 61.54 & 65.00 & 60.94 & 69.06\\
ResGCN-N39 & 52.28 & 53.59 & 50.96 & 49.52 & 45.62 & 53.42\\
ResGCN-N51 & 39.83 & 41.83 & 37.82 & 38.63 & 37.19 & 40.07\\
ResGCN-N57 & 52.63 & 56.86 & 48.40 & 54.30 & 50.62 & 57.98\\
ResGCN-N75 & 41.13 & 44.44 & 37.82 & 37.13 & 38.44 & 35.83\\
FusionGhost-Gait-S & 64.42 & 66.67 & 62.18 & 66.34 & 59.06 & 73.62\\
\bottomrule
\end{tabular}
\end{table}

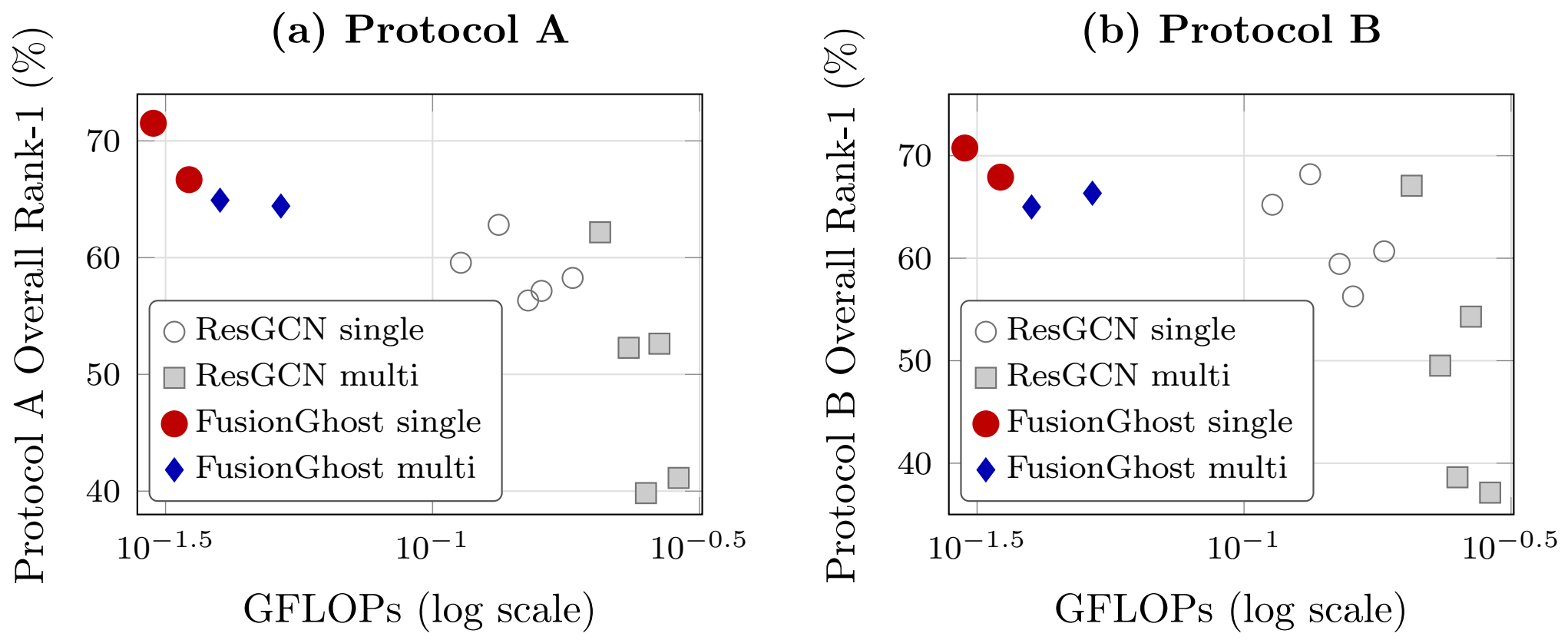
\begin{figure}[!t]
\centering
\begin{minipage}[t]{0.49\textwidth}
\centering
\begin{tikzpicture}
\begin{axis}[
width=0.96\linewidth,
height=5.2cm,
xmode=log,
xlabel={GFLOPs (log scale)},
ylabel={Protocol A Overall Rank-1 (\%)},
xmin=0.028, xmax=0.32,
ymin=38, ymax=74,
grid=both,
grid style={gray!25},
tick label style={font=\scriptsize},
label style={font=\footnotesize},
title style={font=\footnotesize\bfseries},
title={(a) Protocol A},
legend style={
font=\scriptsize,
draw=gray!65!black,
fill=white,
rounded corners=2pt,
at={(0.02,0.03)},
anchor=south west
},
legend cell align=left
]
\addplot[
only marks,
mark=o,
mark size=2.5pt,
draw=black!55,
fill=white
] coordinates {
(0.113,59.56)
(0.133,62.81)
(0.151,56.33)
(0.160,57.15)
(0.183,58.26)
};
\addlegendentry{ResGCN single}

\addplot[
only marks,
mark=square*,
mark size=2.5pt,
draw=black!55,
fill=gray!40
] coordinates {
(0.206,62.17)
(0.233,52.28)
(0.251,39.83)
(0.266,52.63)
(0.289,41.13)
};
\addlegendentry{ResGCN multi}

\addplot[
only marks,
mark=*,
mark size=3.1pt,
color=red!75!black
] coordinates {
(0.03,71.51)
(0.035,66.68)
};
\addlegendentry{FusionGhost single}

\addplot[
only marks,
mark=diamond*,
mark size=2.8pt,
color=blue!70!black
] coordinates {
(0.04,64.92)
(0.052,64.42)
};
\addlegendentry{FusionGhost multi}
\end{axis}
\end{tikzpicture}
\end{minipage}
\hfill
\begin{minipage}[t]{0.49\textwidth}
\centering
\begin{tikzpicture}
\begin{axis}[
width=0.96\linewidth,
height=5.2cm,
xmode=log,
xlabel={GFLOPs (log scale)},
ylabel={Protocol B Overall Rank-1 (\%)},
xmin=0.028, xmax=0.32,
ymin=35, ymax=76,
grid=both,
grid style={gray!25},
tick label style={font=\scriptsize},
label style={font=\footnotesize},
title style={font=\footnotesize\bfseries},
title={(b) Protocol B},
legend style={
font=\scriptsize,
draw=gray!65!black,
fill=white,
rounded corners=2pt,
at={(0.02,0.03)},
anchor=south west
},
legend cell align=left
]
\addplot[
only marks,
mark=o,
mark size=2.5pt,
draw=black!55,
fill=white
] coordinates {
(0.113,65.22)
(0.133,68.19)
(0.151,59.44)
(0.160,56.27)
(0.183,60.67)
};
\addlegendentry{ResGCN single}

\addplot[
only marks,
mark=square*,
mark size=2.5pt,
draw=black!55,
fill=gray!40
] coordinates {
(0.206,67.07)
(0.233,49.52)
(0.251,38.63)
(0.266,54.30)
(0.289,37.13)
};
\addlegendentry{ResGCN multi}

\addplot[
only marks,
mark=*,
mark size=3.1pt,
color=red!75!black
] coordinates {
(0.03,70.75)
(0.035,67.91)
};
\addlegendentry{FusionGhost single}

\addplot[
only marks,
mark=diamond*,
mark size=2.8pt,
color=blue!70!black
] coordinates {
(0.04,65.00)
(0.052,66.34)
};
\addlegendentry{FusionGhost multi}
\end{axis}
\end{tikzpicture}
\end{minipage}
\caption{ArabGait accuracy--efficiency scatter under cross-clothing protocols. FusionGhost points remain in the low-FLOPs region while preserving strong Protocol A/B accuracy. GFLOPs correspond to the same model configurations used in Table~\ref{tab:single_stream} and Table~\ref{tab:multi_stream}.}
\label{fig:arabgait_efficiency}
\end{figure}

FusionGhost-Gait-S is now included directly in
Table~\ref{tab:single_stream} and Table~\ref{tab:multi_stream}. In
single-stream it reaches 66.68\% on Protocol A and 67.91\% on Protocol B.
In multi-stream it reaches 64.42\% on Protocol A and 66.34\% on Protocol B.
The single-stream configuration remains the stronger
cross-clothing operating point, consistent with the earlier observation
that derived streams can amplify pose noise under severe lower-limb
occlusion. Fig.~\ref{fig:arabgait_efficiency} summarizes this
accuracy--efficiency behavior on ArabGait for both Protocol A and
Protocol B.

\subsection{Pose-Degradation Analysis}

Gulf clothing introduces a fundamentally different covariate from the
coat/jacket variations commonly studied in datasets such as CASIA-B.
Kandoora and Abaya garments affect the pipeline at an earlier stage by
reducing joint observability during pose estimation. Loose fabric occludes
lower-limb structure and suppresses motion cues, producing temporally unstable
or low-confidence keypoints. Consequently, the challenge is not only appearance
variation but also degradation of the structural signal available to
skeleton-based models.

As shown in Sec.~\ref{sec:dataset}, this effect is clear in the pose-confidence
profiles (Table~\ref{tab:pose_quality_metrics}, Fig.~\ref{fig:pose_quality_overview}).
The cross-clothing protocols make the same point at the recognition
level. Transferring from casual to Gulf clothing or vice versa is much
harder than the legacy same-clothing evaluation. This pattern indicates
that Gulf clothing should be treated primarily as a pose-estimation
challenge rather than a purely appearance-based covariate, which
motivates explicit cross-clothing evaluation.

\section{Discussion}
\label{sec:discussion}

\subsection{Capacity Sensitivity and Stream Noise Amplification}
\label{sec:capacity}
Tables~\ref{tab:single_stream} and~\ref{tab:multi_stream} reveal that
cross-clothing generalization is capacity-sensitive. In single-stream,
performance improves from N21 to N39 and then declines for larger models.
In multi-stream, larger backbones degrade even more sharply, with N51 and
N75 performing especially poorly on Protocols A and B. This behavior
aligns with the bias--variance perspective, where higher-capacity models
overfit occlusion-corrupted pose patterns instead of learning
clothing-invariant gait dynamics.

This behavior is consistent with the structural occlusion introduced by Gulf clothing,
where unreliable keypoints propagate into derived streams. The generalization
gap is larger in the multi-stream setting, indicating that fusion design
interacts with model capacity under noisy conditions and is not always
beneficial under severe occlusion. For the S-scale baseline, accuracy drops
from 75.35\% in the single-stream original protocol to 67.40\% in the
multi-stream original protocol, then further to 52.28\% and 49.52\% on
the multi-stream cross-clothing Protocols A and B.
We hypothesize that this drop arises from \emph{error propagation} in derived streams.
Multi-stream skeleton pipelines fuse joint coordinates with bone and velocity
representations where bone features are defined as vectors between connected joints,
while velocity streams are constructed by temporal differencing.
Under occlusion, pose estimators produce jittery or mislocalized joints, and these
derived representations inherit and can amplify such errors. This interpretation is
consistent with prior findings that GCN-based skeleton methods are sensitive to
pose-estimation noise~\citep{duan2022revisiting} and that incomplete or noisy
skeletons can significantly degrade recognition performance~\citep{song2020ragcn}.

We empirically validated the noise amplification hypothesis by analyzing the
temporal standard deviation of lower-limb features across streams and conditions
(Table~\ref{tab:stream_variance}). The stream-level statistic is
\begin{equation}
\label{eq:stream_std}
\mathrm{Std}_{s,c}
= \frac{1}{N_{s,c} D_s}
\sum_{n=1}^{N_{s,c}}\sum_{d=1}^{D_s}
\mathrm{std}_t\!\left(F_{n,t,d}^{(s,c)}\right)
\end{equation}
where $s\in\{\text{joints},\text{bones},\text{velocity}\}$ denotes the stream,
$c$ denotes the clothing condition, $N_{s,c}$ is the number of clips in that
group, $D_s$ is feature dimensionality, and
$\mathrm{std}_t(\cdot)$ is the sample standard deviation over time.
Lower-limb variance increases monotonically
from Casual to Gulf with Bag for all three streams, confirming that Gulf clothing
systematically destabilizes lower-body keypoints. Critically, the amplification
is not uniform across streams. Relative to Casual, Gulf clothing increases
lower-limb temporal variance by 10.8\% for joints, 33.6\% for bones, and
21.1\% for velocity. The larger increases in derived streams (bones and
velocity) indicate stronger noise amplification than in joints, so multi-stream
fusion receives comparatively noisier cues.
Ghost modules regularize feature expansion and reduce overfitting to cross-stream noise,
yielding larger gains in this setting.
\begin{table}[!t]
\caption{Lower-limb temporal standard deviation per stream across clothing conditions. The final column reports percentage increase for Gulf relative to Casual clothing.}
\label{tab:stream_variance}
\centering
\scriptsize
\setlength{\tabcolsep}{3pt}
\begin{tabular}{lccccc}
\toprule
\textbf{Stream} &
\shortstack{\textbf{Casual}\\\textbf{Clothing}} &
\shortstack{\textbf{Casual Clothing}\\\textbf{ with Bag}} &
\shortstack{\textbf{Gulf}\\\textbf{Clothing}} &
\shortstack{\textbf{Gulf}\\\textbf{Clothing with Bag}} &
\shortstack{\textbf{Gulf}\\\textbf{vs Casual (\%)}} \\
\midrule
Joints   & 132.27 & 139.15 & 146.55 & 154.66 & $+10.8$\% \\
Bones    &   9.89 &  11.28 &  13.21 &  14.50 & $+33.6$\% \\
Velocity & 108.54 & 120.65 & 131.39 & 147.57 & $+21.1$\% \\
\bottomrule
\end{tabular}
\end{table}

\subsection{Ghost Factorization as Capacity Control Under Occlusion}
\label{sec:ghost}

The degradation patterns in Sec.~\ref{sec:capacity} indicate that
over-parameterized models and non-adaptive fusion amplify noise under
occlusion, particularly when multiple derived streams are fused. Ghost
modules provide a structural countermeasure by reducing intrinsic feature
channels and generating the remaining features through depthwise
operations~\citep{han2020ghostnet}. This constrained parameterization limits
feature expansion and behaves as an implicit regularizer, consistent with
compressed or factorized models where reduced capacity can improve
robustness under noisy supervision~\citep{furlanello2018bornagain}.

Under Gulf-clothing occlusion, this regularization reduces memorization of
unstable pose jitter and encourages reliance on more persistent motion cues.
Prior work on skeleton-based models reports similar sensitivity to noisy or
incomplete poses~\citep{song2020stronger}, and within this setting Ghost
factorization promotes feature reuse instead of uncontrolled channel growth.
The effect is therefore not only parameter reduction, but capacity control in
the presence of stream-dependent noise.

FusionGhost-Gait operationalizes this capacity-control principle with a fused
spatiotemporal block that removes the heavy dedicated TCN branch and injects
temporal context through compact temporal Ghost kernels while preserving
lightweight temporal downsampling. This design keeps temporal modeling
explicit but reduces redundant channel mixing under pose-noisy inputs.

Empirically, FusionGhost-Gait-S supports the same interpretation. In
single-stream, the final design reaches 74.96\% under the legacy
protocol and remains at 66.68\% and 67.91\% under Protocols A and B.
In multi-stream, it reaches 71.13\%, 64.42\%, and 66.34\% under the
same three evaluations while using only 133,674 parameters and
0.052 GFLOPs. Together, these findings suggest reduced harmful
cross-stream co-adaptation rather than mere lightweight substitution.
This trend is also aligned with results in compression and distillation,
where constrained students can outperform larger teachers when data are
noisy or limited~\citep{hinton2015distilling}.

\section{Limitations}

This study has limitations that affect generalizability. ArabGait is indoor and relatively
small (45 subjects, 10 test identities in one subject-disjoint split), so small accuracy
differences may be split-sensitive. ArabGait evaluation is deliberately
identical-view because the dataset is intended to isolate
clothing-induced pose degradation rather than to serve as a full
cross-view benchmark. Public grounding is therefore provided separately
through CASIA-B, but broader cross-dataset and cross-view validation
remains future work. The analysis uses one pose estimator (HRNet-W48),
one Ghost ratio ($s=2$), and reports parameters/FLOPs rather than
wall-clock latency, so estimator dependence, ratio sensitivity, and
runtime behavior remain under-characterized.

Future work will report repeated-split or cross-validation results with uncertainty estimates,
add cross-view and cross-dataset benchmarks, evaluate a second pose pipeline, perform
Ghost-ratio ablations, and provide runtime measurements for key variants.
We will also compare against stronger external baselines and investigate
3D pose lifting to reduce occlusion sensitivity.

\section{Conclusion}
\label{sec:conclusion}

This paper introduced ArabGait, a new multi-view skeleton gait dataset
for severe Gulf-clothing occlusion (Kandoora/Abaya), and showed that
the dominant failure mode is upstream pose degradation rather than
appearance variation alone. Across conditions, the analysis highlighted
two core mechanisms. Multi-stream fusion can amplify occlusion-induced
noise, and model capacity must be controlled to avoid overfitting under
noisy pose inputs.

FusionGhost-Gait is the proposed architectural solution developed from
these findings. The fused design represents temporal dependency through
lightweight Ghost branches inside the spatial block and preserves a
strong efficiency--accuracy operating point.
Public CASIA-B cross-view results provide external comparability, while
ArabGait cross-clothing experiments show why this regime matters.
Gulf clothing degrades pose quality, cross-clothing transfer exposes that
failure more directly than the legacy protocol, and capacity-controlled
fusion is better suited to this pose-noisy setting than naive scaling.


\section*{CRediT authorship contribution statement}
\textbf{Seba Al Mokdad:} Writing -- original draft, Investigation, Conceptualization, Methodology, Validation, Visualization, Data curation. \textbf{Tamer Shanableh:} Funding acquisition, Supervision, Project administration, Writing -- review \& editing, Conceptualization. \textbf{Reham Aburas:} Funding acquisition, Supervision, Writing -- review \& editing, Resources.

\section*{Funding}
This work was supported by the American University of Sharjah Faculty Research Grant (Grant No. FRG25-C-E51).

\section*{Declaration of competing interest}
The authors declare that they have no known competing financial interests or personal relationships that could have appeared to influence the work reported in this paper.

\section*{Data availability}
The ArabGait dataset will be made available upon acceptance of this paper.

\section*{Project repository}
\url{https://github.com/seba20-0/FusionGhostGait}

\section*{Ethics approval and informed consent}
This study was approved by the American University of Sharjah Institutional Review Board under Protocol \#24-039. Written informed consent was obtained from all participants before data collection.

\section*{Declaration of generative AI and AI-assisted technologies in the manuscript preparation process}
During the preparation of this work, the authors used OpenAI Codex to assist with LaTeX reformatting and language presentation. After using this tool, the authors reviewed and edited the content as needed and take full responsibility for the content of the publication.

\bibliographystyle{elsarticle-harv}
\bibliography{arabgait}
\end{document}